\documentclass[journal]{IEEEtran}
\usepackage{cite}
\usepackage{graphicx}
\usepackage{booktabs}
\usepackage{tikz}
\usepackage{eurosym}
\usepackage{alphalph,etoolbox}
\usepackage{url}
\usepackage{epstopdf}
\usepackage{amsfonts,amsmath,amsthm,amssymb}
\usepackage{subfigure}
\usepackage{stfloats}
\usepackage{eurosym}
\usepackage{enumerate}
\usepackage{color}
\usepackage{arydshln}
\usepackage{longtable}
\usepackage{mathrsfs}
\usepackage{bm}
\usepackage[T1]{fontenc}
\usepackage[latin1]{inputenc}
\usepackage[shortlabels]{enumitem}
\usepackage{tikz}
\usepackage{footnote}
\makesavenoteenv{minipage}
\makesavenoteenv{enumerate}
\usepackage{endnotes}
\usetikzlibrary{arrows}

\usepackage{csquotes}

\usepackage{psfrag}

\patchcmd{\subequations}{\alph{equation}}{\alphalph{\value{equation}}}{}{}

\newtheorem*{remark}{Remark}

\newtheorem{assump}{Assumption}

\include{def}

\begin{document}

\graphicspath{{Figures/}}

\title{Optimization- and AI-based approaches to academic quality quantification
for transparent academic recruitment: part 1-model development}


\author{Ercan Atam

\thanks{
Ercan Atam is with Institute for Data Science \& Artificial Intelligence
of Bo\u{g}azi\c{c}i University, Istanbul, Turkey.
E-mail: ercan.atam@boun.edu.tr 
}}


\maketitle

\begin{abstract}
For fair academic recruitment at universities and research institutions,
determination of the right measure based on globally accepted academic quality features is a highly delicate, challenging, but quite important problem to be addressed.
In a series of two papers, we consider the modeling
part for academic quality quantification in the first paper, in this paper, and the case studies part in the second paper. 
For academic quality quantification modeling, we develop two computational frameworks which can be used to construct 
a decision-support tool:
(i) an optimization-based  framework  and 
(ii) a Siamese network (a type of artificial neural network)-based framework. 
The output of both models is a single index called Academic Quality Index (AQI)
which is a measure of the overall academic quality.
The data of academics from first-class and average-class world universities,
based on Times Higher Education World University Rankings
and QS World University Rankings, are assumed as the reference data for tuning model parameters.


\end{abstract}

\IEEEpeerreviewmaketitle

\section{Introduction}

Universities play a  central role in higher education, community service and advancing science \& technology. For all nations, self-sufficiency, self-sustainability, leading science and technology,
and remaining in global competition are possible only by owning high-quality universities. Two  key constituents of a good university
are high-quality students and academics. Proper selection of the best academics for faculty hiring 
is not an easy task, especially when universities receive lots of
successful applicants with different quality features for an advertised position.
In such cases, in addition to the challenges of  proper assessment, 
the selection process itself takes a long time and effort.

Another related issue in academic recruitment is the lack of a transparent and auditable evaluation system in many universities across the globe. Today, the widely used procedure 
for academic personnel recruitment
is selection of expert committees for assessment of applicants.
However, most of the time the members have different
preferences on which candidate to choose
and  selection is  done
by the maximum vote method where the candidate receiving
the highest number of votes is selected.
In theory this procedure seems okay.
The main problem here is on the ``subjectiveness''
of the jury members in the selection phase.
Academic quality can be characterized by
a large number of features (see Table~\ref{table:Features characterizing academic quality}
in Section~\ref{sec:Input features}
for a long list of  some of these features that we considered important), and  for each member
how much each feature is important may change dramatically. 
On top of this, academic inbreeding, backing through networks or connections, preferring ideologically-consistent candidates or candidates with common
interests, especially
at non-institutionalized and corrupted universities,
are the de facto realities and can influence the voting behaviour considerably. All these facts together question the fairness of
the current academic recruitment process in many universities
and research institutions.
Hence, it is natural to ask the following questions: Is it possible
to define somehow academic quality? Can we quantify it and then develop
a tool to provide help in decision making during the academic recruitment process?

In this paper, we  concern with the above questions and give an attempt to address the
problem of academic quality quantification
by defining and predicting an index called  Academic
Quality Index (AQI),
which has not been considered before
in the literature to the
best knowledge of author,
but it can be quite important 
for the process of academic personnel selection.
We contribute to the literature on
academic quality quantification by presenting two different approaches
for AQI prediction: an optimization-based
and a Siamese network-based prediction method.
The common goal of both methods is to determine a model
which takes as input a set of features characterizing academic quality
and output an index
which measures the overall academic quality.
The model weights are optimally tuned using data of academics from 
first-class world universities (called ``positive class'') where academics
in general have high-quality academic features, and data
of academics from other universities (called ``negative class'') where
academics in general have relatively lower quality profiles.

In the optimization-based method, linear and quadratic regression
models based on academic quality features are constructed,
and optimization is used to  simultaneously minimize dissimilarity between
academics from the positive class and maximize  dissimilarity between the positive and negative classes. The main advantage of optimization-based approach is that relevant constraints (such as enforcing the condition that on average
academics in the positive class have higher AQIs compared to those
in the negative class, and including ranking of model weights)
can  be easily included. The main disadvantage, in practice, is the difficulty of
integrating complex models (such as deep ANN models) into optimization.

The alternative approach, Siamese network-based prediction, has
the opposite features: allows complex model architectures, but it is
hard to include constraints. As a result, the two methods can be regarded
as complementary approaches. A Siamese network \cite{Bromley_et_al_1998} is
an ANN consisting of identical subnetworks (sharing the same weight and structure),
which is used for similarity learning. In the literature, there are many 
different prediction or assessment applications using varieties of Siamese networks: for
example,
in \cite{Annaland_et_al_2020} for cherry quality prediction, in \cite{Jain_et_al_2020}
for action quality assessment (determining how well an
action was performed), in \cite{Liu_et_al_2017} for image quality assessment, 
in \cite{Chang_et_al_2020} for knee pain level prediction from MR scans, 
in \cite{Neculoiu_et_al_2016} for text similarity assessment,
in \cite{Wang_et_al_2018} for road detection from the perspective of moving vehicles,
in \cite{Zhao_et_al_2019} for software defect prediction,
and in \cite{Ramachandra_et_al_2020} for video anomaly prediction and detection  (localizing anomalies spatially and temporally)  where anomalies refer to unusual activities. In this paper, we will use it for another application: AQI prediction. The main advantage of Siamese networks compared to other
ANN architectures or learning methods is that they
have a good ability of learning  in cases
(i) very little training data are available, or
(ii) similar and dissimilar classes are imbalanced
\cite{Koch_et_al_2015, Neculoiu_et_al_2016, Wang_et_al_2018}.

The rest of the paper is structured as follows.
In Section~\ref{sec:Input features},
creation of input features characterizing academic quality
is presented.
Next, we discuss creation of reference input data for
model tuning in Section~\ref{sec:Reference data creation for model tuning}.
Optimization-based framework for academic quality quantification is given 
in Section~\ref{sec:Optimization-based modeling for academic quality quantification}
and ANN-based one in Section~\ref{sec:ANN-based modeling for academic quality quantification}.
A guideline flowchart summarizing model development for academic quality quantification and the use
of the resulting models is given in Section~\ref{sec:A guideline based on the developed frameworks for academic recruitment}.
Finally, we conclude and list some future research directions
in Section~\ref{sec:Conclusions}.


\section{Input features}
\label{sec:Input features}

Given an academic, there are many features which can reflect
his/her academic quality.
A long list of main inputs to construct features reflecting academic quality is given in
Table~\ref{table:Inputs for feature construction}.
We observe that the given input set tries to encompass two main 
academic quality dimensions: (i) ``research capacity" (mainly represented
by the number of publications at top journals, the number of citations and related indices,
the number of patents, the  number of prestigious  national or   international projects as principal investigator, the number of awards and recognized work, and the number of supervised PhD students),  (ii) ``knowledge capacity"
(mainly represented by the national or international ranking of
the university where BSc and PhD degrees were obtained, the grade point averages (GPAs), the number of books from a prestigious publisher,
and the number of different undergraduate/graduate courses that can be given).
Note that research capacity and knowledge capacity do not necessarily imply each other;
they may be correlated in some cases; and both are extremely essential for academics to be hired as faculty members. We must stress that 
although 
the list of inputs in Table~\ref{table:Inputs for feature construction} related to academic quality
contains more or less globally well-accepted inputs, still they are based on author's subjective opinion,
and hence, it is possible to include
additional inputs and/or construct a se of different inputs.

\begin{table}[h!]
\centering
\caption{Inputs for feature construction}
\label{table:Inputs for feature construction}
 \begin{tabular}{|p{1.8cm}||p{6cm}|}
   \hline
   \bf{Input} &  \bf{Description}   \\ \hline \hline
    $n_{q_1}$ & number of  SCI-SCIE papers in the ${Q_1}$ quartile  \\  \hline
    $n_{q_2}$ & number of  SCI-SCIE papers in the ${Q_2}$ quartile \\  \hline
    $n_{q_3}$ & number of  SCI-SCIE papers in the  ${Q_3}$ quartile \\  \hline
    $n_{q_4}$ & number of  SCI-SCIE papers in the ${Q_4}$ quartile \\  \hline
    $n_\text{$q_1$-ave-auth}$ &  average number of authors for SCI-SCIE papers  in the ${Q_1}$ quartile \\  \hline
    $n_\text{$q_2$-ave-auth}$ &  average number of authors for SCI-SCIE papers  in the ${Q_2}$ quartile \\  \hline
    $n_\text{$q_3$-ave-auth}$ &  average number of authors for SCI-SCIE papers  in the ${Q_3}$ quartile \\  \hline
    $n_\text{$q_4$-ave-auth}$ &  average number of authors for SCI-SCIE papers  in the ${Q_4}$ quartile \\  \hline
    $n_\text{q1-fa}$ & number of  SCI-SCIE papers in the ${Q_1}$ quartile as a first author  \\  \hline
    $n_{\text{conf}}$ & number of prestigious conference papers in the field    \\  \hline
    $n_{\text{conf-ave-auth}}$ & average number of authors for prestigious conference papers in the field   \\  \hline
    $n_{\text{book}}$ & number of books from a prestigious publisher\\  \hline
    $n_{\text{book-ave-auth}}$ &  average number of authors for the books from a prestigious publisher\\  \hline
    $n_{\text{book-chp}}$ & number of book chapters for books from a prestigious publisher\\  \hline
    $n_{\text{book-chp--ave-auth}}$ &  average number of authors for the book chapters from a prestigious publisher\\  \hline
    $n_{\text{cit}}$ & number of citations \\ \hline
    $h_\text{ind}$ & h-index \\  \hline
    $i10_\text{ind}$ & i10-index \\  \hline
    $n_{\text{pat}}$ & number of patents \\  \hline
    $n_{\text{pat-ave-auth}}$ &  average number of authors for the patents\\  \hline
    $n_{\text{prj}}$ & number of prestigious national or international projects as  principal 
    investigator (PI) \\  \hline
    $n_{\text{award-recog work}}$ & number of awards and recognized work \\  \hline
    $n_{\text{MS-stud}}$ & number of supervised MS students \\  \hline
    $n_{\text{PhD-stud}}$ & number of supervised PhD students \\  \hline
    $t_{\text{res}}$ & time from start of PhD to present for research \\  \hline
    $t_{\text{res}}^{'}$ & time from PhD graduation to present for research \\  \hline
    $r_{\text{nat-BS}}$ & national ranking of university where BS degree was obtained, if applies  \\  \hline
    $r_{\text{nat-PhD}}$ & national ranking of university where PhD degree was obtained, if applies  \\  \hline
    $r_{\text{inat-BS}}$ & international ranking of university where BS degree was obtained, if applies  \\  \hline
    $r_{\text{inat-PhD}}$ & international ranking of university where PhD degree was obtained, if applies  \\  \hline
    $\text{GPA}_u$ & undergraduate GPA  \\ \hline 
    $\text{GPA}_g$ & graduate GPA  \\ \hline 
    $n_{\text{course-u}}$ & number of different undergraduate courses given or that can be given \\  \hline
    $n_{\text{course-g}}$ & number of different graduate courses given  or that can be given \\  \hline
 \end{tabular}
\end{table}

The constructed features from  the input data in Table~\ref{table:Inputs for feature construction} are given in Table~\ref{table:Features characterizing academic quality}
where
\begin{align*}
& \displaystyle \bar{n}_{q_i} \triangleq \frac{n_{q_i}}{n_\text{qi-ave-auth} \times t_{\text{res}}},\, i=1,\cdots,4,\, \bar{n}_{\text{conf}}  \triangleq \frac{n_{\text{conf}}}{n_{\text{conf-ave-auth}} \times t_{\text{res}}},\quad \\
& \bar{n}_{\text{book}} \triangleq \frac{n_{\text{book}}}{n_{\text{book-ave-auth}} \times t_{\text{res}}},\quad 
\bar{n}_{\text{book-chp}} \triangleq \frac{n_{\text{book-chp}}}{n_{\text{book-chp-ave-auth}} \times t_{\text{res}}},\quad \\
&\bar{n}_{\text{pat}} \triangleq \frac{n_{\text{pat}}}{n_{\text{pat-ave-auth}} \times t_{\text{res}}},\quad
\bar{n}_{\text{prj}} \triangleq \frac{n_{\text{prj}}}{t_{\text{res}}},\quad \bar{n}_{\text{cit}} \triangleq \frac{n_{\text{cit}}}{t_{\text{res}}} \\
& \bar{h}_{\text{ind}} \triangleq \frac{h_{\text{ind}}}{t_{\text{res}}},\quad 
\overline{i10}_{\text{ind}} \triangleq \frac{i10_{\text{ind}}}{t_{\text{res}}},\, 
\bar{n}_{\text{award-recog-work}} \triangleq \frac{n_{\text{award-recog-work}}}{t_{\text{res}}},\quad \\
&  \bar{n}_{\text{MS-stud}}  \triangleq \frac{n_{\text{MS-stud}}}{t_{\text{res}}^{'}},\quad  \bar{n}_{\text{PhD-stud}} \triangleq \frac{n_{\text{PhD-stud}}}{t_{\text{res}}^{'}}
\end{align*}

The fourth column in Table~\ref{table:Features characterizing academic quality}
includes a subjective (but reasonable) ordering of the importance of each feature,
which will be integrated into the
optimization-based academic quality quantification models developed
in Section~\ref{sec:Optimization-based modeling for academic quality quantification}. The proposed feature ranking is just a suggestion, and if desired, it can be changed to any other  preferred ranking.

\begin{table} 
\centering
\caption{Features characterizing academic quality}
\label{table:Features characterizing academic quality}
\begin{tabular}{|p{1.6cm}||p{0.5cm}|p{4cm}|l|}
  \hline
  \textbf{Feature} & \textbf{Rep.} & \textbf{Description}  & \textbf{Rank} \\ \hline
  $\bar{n}_{q_1}$ & $x_1$ &  Normalized number of $Q_1$ papers & 1 \\  \hline
  $\bar{h}_{\text{ind}}$ &  $x_2$ &  Normalized h-index  & 2 \\ \hline
  $\bar{n}_{\text{cit}}$ & $x_3$ & Normalized number of citations & 3 \\ \hline
  $\overline{i10}_{\text{ind}}$ & $x_4$  & Normalized i10-index & 4 \\ \hline
  $\bar{n}_{\text{book}}$ & $x_5$ & Normalized number of books from a prestigious publisher & 5 \\  \hline
  $\bar{n}_{\text{award-recog work}}$ & $x_6$ & Normalized number of awards and recognized work & 6 \\  \hline
  $r_{\text{inat-PhD}}$ & $x_7$ & International ranking of university where PhD was obtained, if applies & 7 \\  \hline
  $\bar{n}_{\text{pat}}$ & $x_8$ & Normalized number of patents & 8 \\  \hline
  $\bar{n}_{\text{prj}}$ & $x_9$ & Normalized number of prestigious national or international projects as PI & 9 \\ \hline
  $\bar{n}_{\text{PhD-stud}}$ & $x_{10}$ & Normalized number of PhD students & 10 \\ \hline
  $\bar{n}_{q_2}$ & $x_{11}$ & Normalized number of $Q_2$ papers & 11 \\ \hline
  $r_{\text{nat-PhD}}$ & $x_{12}$ &  National ranking of university where PhD degree was obtained, if applies & 12 \\  \hline
  $r_{\text{inat-BS}}$ & $x_{13}$ &  International ranking of university where BS degree was obtained, if applies & 13 \\  \hline
  $r_{\text{nat-BS}}$ & $x_{14}$ &  National ranking of university where BS degree degree was obtained, if applies & 14 \\  \hline
  $\bar{n}_{\text{MS-stud}}$ & $x_{15}$ &  Normalized number of MS students & 15 \\ \hline
  $\text{GPA}_g$ & $x_{16}$ & Graduate GPA &  16 \\ \hline 
  $\text{GPA}_u$ & $x_{17}$ &  Undergraduate GPA & 17  \\ \hline 
  $\bar{n}_{q_3}$ & $x_{18}$ & Normalized number of $Q_3$ papers & 18 \\  \hline
  $\bar{n}_{q_4}$ & $x_{19}$ &  Normalized number of $Q_4$ papers & 19 \\ \hline
  $\bar{n}_{\text{book-chap}}$ & $x_{20}$  & Normalized number of book chapters & 20 \\  \hline
  $\bar{n}_{\text{conf}}$ & $x_{21}$  &  Normalized number of prestigious conference papers & 21 \\
  \hline
\end{tabular}
\end{table}

\begin{remark}
The construction of features involving studies where multiple authors
collaborate can be modified
to take into account the author order. In general, the first author does most of the work
and the listing of authors is in order of decreasing contribution.
\end{remark}




\section{Reference data creation for model tuning}
\label{sec:Reference data creation for model tuning}

In addition to  model development for academic quality
quantification, another important step  is creation of reference input data for model tuning,
which is probably  open to discussion. To that end, we propose the following assumptions.

\begin{assump} \label{similarity_assumption}
For a given academic level (Assist. Prof., Assoc. Prof., or Prof.), an academic field and research type (theoretical research versus applied research), the academics  at top 20 world universities based on
a well-known and widely accepted ranking system (such as Times Higher Education World University Rankings)
can be assumed to have ``similar" academic quality, and their input data can be taken as reference data for model tuning.
\end{assump}

\begin{remark} Many academics have mixed research interests:
a mix of theoretical and applied research. Handling of such cases
will be discussed in Section~\ref{sec:A guideline based on the developed frameworks for academic recruitment}.
\end{remark}

\begin{assump} \label{dissimilarity_assumption}
For a given academic level (Assist. Prof., Assoc. Prof., or Prof.), an academic field and research type (theoretical research versus applied research),
the top 20 world universities based on a well-known and widely accepted ranking system (such as Times Higher Education World University Rankings)
can be assumed, in general, to have considerably  better
academics (in the sense of better academic quality
features) than those in average-ranked universities.
In model tuning, we will use data of academics from average-ranked
universities as well.
\end{assump}

Note that these two assumptions can be
seen controversial.
However, still we believe that they are
reasonable assumptions on which majority of academics
would agree.

\section{Optimization-based modeling for academic quality quantification}
\label{sec:Optimization-based modeling for academic quality quantification}

\subsection{Parameterized regression models}
\label{subsec:Parameterized regression models}

In this section, we will present two regression models to be used 
for academic quality prediction. Before that, let $\bar{x}$ denote the normalized feature vector obtained by proper normalization of each component of feature vector $x$ (see Table \ref{table:Features characterizing academic quality})
so that $\bar{x}_i \in [0, 1]$. Now, a generic regression model can be represented 
by $f(\bar{x},w)$ where $w$ is
weight parameter vector to be tuned optimally using numerical optimization,
which will be used to include the relevant constraints as well.

The academic quality index (AQI) is an index in the range $[0,\,100]$,
and for an academic $i$ with normalized
feature vector $\bar{x}^i$, it is defined as
\begin{equation} \label{AQI_def}
\text{AQI}(w,\bar{x}^i)\triangleq \displaystyle 100 \times f(w,\bar{x}^i)
\end{equation}
where $f(w,\bar{x}^i)$ will be constrained so that $f(w,\bar{x}^i) \in [0,\,1]$. 



\subsubsection{$\text{M}_\text{1}$: Linear Regression Model}
\label{subsubsec:M1}

The first regression model we consider is
the linear regression model given in \eqref{M1}.
This model is the simplest one and it has $21$ parameters
($w=\alpha$) to be tuned.

\subsubsection{$\text{M}_\text{2}$: Quadratic Regression Model}
\label{subsubsec:M1}

The second regression model we consider is
the quadratic regression model given in \eqref{M2}.
The quadratic regression model is relatively complex, and has $21+21+\binom{21}{2}=252$ parameters
($w=(\alpha, \beta, \theta)^T$) to be tuned.

\begin{figure*}[ht!]
\hrule
\vspace{0.1cm}
$\text{M1: linear regression model}$:
\vspace{0.1cm}
\hrule
\begin{align} \label{M1}
f(\bar{x},\alpha)= \sum_{i=1}^{21}\alpha_i\bar{x}_i, \quad \bar{x}=[\bar{x}_1,\,\bar{x}_2,\cdots, \bar{x}_{21}]^T, \quad \bar{x}_i \in [0, 1]
\end{align}
\hrule
\end{figure*}

\begin{figure*}[ht!]
\hrule
\vspace{0.1cm}
$\text{M2: quadratic regression model}$:
\vspace{0.1cm}
\hrule
\begin{align} \label{M2}
f(\bar{x},\alpha,\beta,\theta)= \sum_{i=1}^{21}\alpha_i\bar{x}_i+
\sum_{i=1}^{21}\beta_i\bar{x}_i^2+
\sum_{\substack{i, j \in \{1,\cdots,21\} \\ i<j}  }\theta_{ij}\bar{x}_i\bar{x}_j
, \quad \bar{x}=[\bar{x}_1,\,\bar{x}_2,\cdots, \bar{x}_{21}]^T, \quad \bar{x}_i \in [0, 1]
\end{align}
\hrule
\end{figure*}

\subsection{Optimization formulation}
\label{subsection:Optimization formulation}

In the optimization formulation, the key and challenging
step is determination of the cost function.
It is clear that comparing two academics
$i$ and $j$ with quality features $\bar{x}^i$ and $\bar{x}^j$, respectively, will
always involve some subjectiveness. As a result, 
the only way to get out of this always-open-to-discussion issue
is to construct a reasonable and widely acceptable cost function.
In the construction of the cost function, we make use of
Assumptions~\ref{similarity_assumption}and~\ref{dissimilarity_assumption}. 

The multi-objective optimization problem valid
for both models $M_1$ and $M_2$
is given in \eqref{opt_problem}.
Its cost function is totally based on the two assumptions.
The input feature pairs used in tuning
model parameters belong to two
sets $\mathcal{S}_p\times \mathcal{S}_p$
and $\mathcal{S}_p\times \mathcal{S}_n$
where $\mathcal{S}_p$ is the set of academics in top 20 world universities with high quality (``positive'')
features, and $\mathcal{S}_n$ is the set of academics in some other average-ranked universities with relatively average quality (``negative'') features.
\textit{I.e.}, we use input data from similar and dissimilar classes,
and use of such data is essential in determining optimal weights of features in order to see which features most contribute to academic quality. 
The cost function is multi-objective (parameterized through $\gamma$) whose
first part measures
the dissimilarity of AQIs of academics in the set $\mathcal{S}_p$ and its second part the dissimilarity of the AQIs of the academics in the two dissimilar classes 
$\mathcal{S}_p$ and $\mathcal{S}_n$
\textit{I.e.}, the cost function is
designed in such a way that it determines
optimal weights so that  simultaneously  (i) AQIs of academics in $\mathcal{S}_p$ 
have minimum dissimilarity and (ii) AQIs of academics in dissimilar classes $\mathcal{S}_p$ and $\mathcal{S}_n$ have
maximum dissimilarity (thanks to the minus sign in front of the second
summation).

\begin{figure*}[ht!]
\hrule
\vspace{0.1cm}
Optimization problem (for $M_1$ \& $M_2$):
\vspace{0.1cm}
\hrule
\begin{subequations} \label{opt_problem}
\begin{eqnarray}
& \hspace{-1cm} \displaystyle \min_{w} \Bigg\{\sum_{(i,j) \in  \mathcal{S}_p \times \mathcal{S}_p} \big(\text{AQI}(w,\bar{x}^i)-\text{AQI}(w,\bar{x}^j)\big)^2 -
\gamma \times \sum_{(i,j) \in  \mathcal{S}_p \times \mathcal{S}_n} \big(\text{AQI}(w,\bar{x}^i)-\text{AQI}(w,\bar{x}^j)\big)^2 \Bigg\}&  \label{obj_func}\\
&\text{AQI}(w,\bar{x}^i)= \displaystyle 100\times f(w,\bar{x}^i), \quad i \in \mathcal{S}_p \cup \mathcal{S}_n   & \label{AQI_def_constraint}  \\
& \displaystyle \frac{1}{|\mathcal{S}_p|}\sum_{i\in\mathcal{S}_p}\text{AQI}(w,\bar{x}^i) \ge  \frac{1}{|\mathcal{S}_n|}\sum_{i\in\mathcal{S}_n}\text{AQI}(w,\bar{x}^i) \label{mean_S_p_greater_mean_Sn_constraint} \\
& w(k) \ge w(l) \text{ if } \text{rank}(k) \le  \text{rank}(l),\quad 1 \le k,l\le n_{w},\, k\ne l 
\quad \text{(for $M_1$)} \label{weight_rank_constraint} \\
& 0 \le r_\text{min}(k) \le  w(k) \le r_\text{max}(k) \le 1, \quad 1 \le k\le n_w \label{w_max_value_constraint} \\
&  \displaystyle  \sum_{k=1}^{n_w}w_k=1 & \label{w_convex_constraint} 
 \end{eqnarray}
\end{subequations}
\hrule
\end{figure*}
As to the constraints, the constraint~\eqref{AQI_def_constraint} is the definition of $\text{AQI}(w,\bar{x}^i)$;
\eqref{mean_S_p_greater_mean_Sn_constraint} indicates that the mean AQI 
of academics in the positive class is greater than or equal to
the mean AQI of academics in the negative class;
\eqref{weight_rank_constraint}, which is valid
only for model $M_1$, forces weights of features with better rankings to be higher;
\eqref{w_max_value_constraint} limits the values each weight can take;
and convex combination requirement for $w$ is given by the
constraints~\eqref{w_convex_constraint}.
One must pay attention in specifying $r_{\text{min}}$
and  $r_{\text{max}}$ in \eqref{w_max_value_constraint} in
order to make sure that constraint~\eqref{w_convex_constraint} is feasible.
Note that the combination of the cost function with
constraint \eqref{mean_S_p_greater_mean_Sn_constraint} achieves
the following:
(i)  places the cluster of AQI of academics in $\mathcal{S}_p$
to the right of  cluster of AQI of academics in $\mathcal{S}_n$ (albeit to some possible overlapping),
(ii) forces the  elements in the cluster of AQI of academics in $\mathcal{S}_p$ to have minimum variance between them, and
(iii) increases the distance between positive and negative classes.

Finally, note that an important structural property to be
satisfied by $f(\bar{x},w)$ is that it must be a monotonically increasing function when a feature value is replaced with a better one.
It can be seen easily that the model structures $M_1$\& $M_2$ together with the associated optimization formulation in \eqref{opt_problem} guarantee this property since
all features and their weights are positive.

\section{ANN-based modeling for academic quality quantification}
\label{sec:ANN-based modeling for academic quality quantification}

In this section, we will use an ANN approach based on the Siamese network \cite{Bromley_et_al_1998}
as an alternative approach for academic quality quantification.
An ANN model can involve complex nonlinear functions and multi-layers, and hence has the potential of performing better than the linear and quadratic regression models given in the previous section.
We will consider two types of Siamese networks corresponding to two cost functions.

\subsection{Siamese network with contrastive loss}
\label{subsection:Siamese network with contrastive loss}

\begin{figure*}[ht!]
\centering
\includegraphics[scale=0.4]{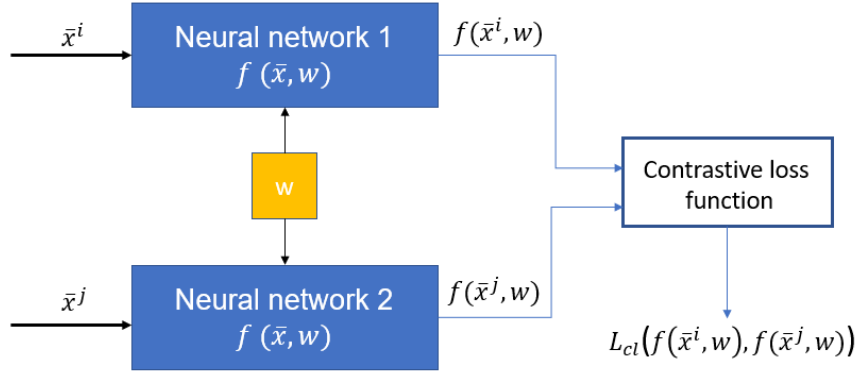}
\caption{Siamese network with contrastive los function.}
\label{fig:Unconstrained Siamese network with contrastive loss function}
\end{figure*}

The Siamese network with contrastive loss function  is
given in Figure~\ref{fig:Unconstrained Siamese network with contrastive loss function}.
The contrastive loss function is defined as 
\begin{align} \label{s_contrastive loss function}
L_{cl}\big(f(\bar{x}^i,w),f(\bar{x}^j,w)\big)  \triangleq   y_L\|f(\bar{x}^i,w)-f(\bar{x}^j,w)\|^2 + \nonumber\\
 (1-y_L) \max\big(m-\|f(\bar{x}^i,w)-f(\bar{x}^j,w)\|, 0\big)^2
\end{align}
where  $m$ is a ``margin'' (user-defined parameter) used to help us in  pushing dissimilar
academics (in the sense of academic quality) apart, and $y_L$ is
a binary variable labeling similar ($y_L$=1) or dissimilar
($y_L=0$)  pairs of academics. The idea behind this type of Siamese network is as follows.
When the input features $\bar{x}^i$ and $\bar{x}^j$ are from the same class,
then $y_L=1$ and we have $L_{cl}\big(f(\bar{x}^i,w),f(\bar{x}^j,w)\big)= \|f(\bar{x}^i,w)-f(\bar{x}^j,w)\|^2$. In this case, in model tuning the weights $w$ are chosen to minimize
this function (to minimize the dissimilarity in the same class). On the other hand, when the input features $\bar{x}^i$ and $\bar{x}^j$ are from the different classes, then $y_L=0$ and we have $L_{cl}\big(f(\bar{x}^i,w),f(\bar{x}^j,w)\big)=
\max\big(m-\|f(\bar{x}^i,w)-f(\bar{x}^j,w)\|, 0\big)^2$. Now, in model tuning the weights $w$
are chosen in order to push $f(\bar{x}^i,w)$ and $f(\bar{x}^j,w)$ at least $m$ units apart.

\subsection{Siamese network with triplet loss}
\label{subsection:Siamese network with triplet losss loss}

\begin{figure*}[ht!]
\centering
\includegraphics[scale=0.4]{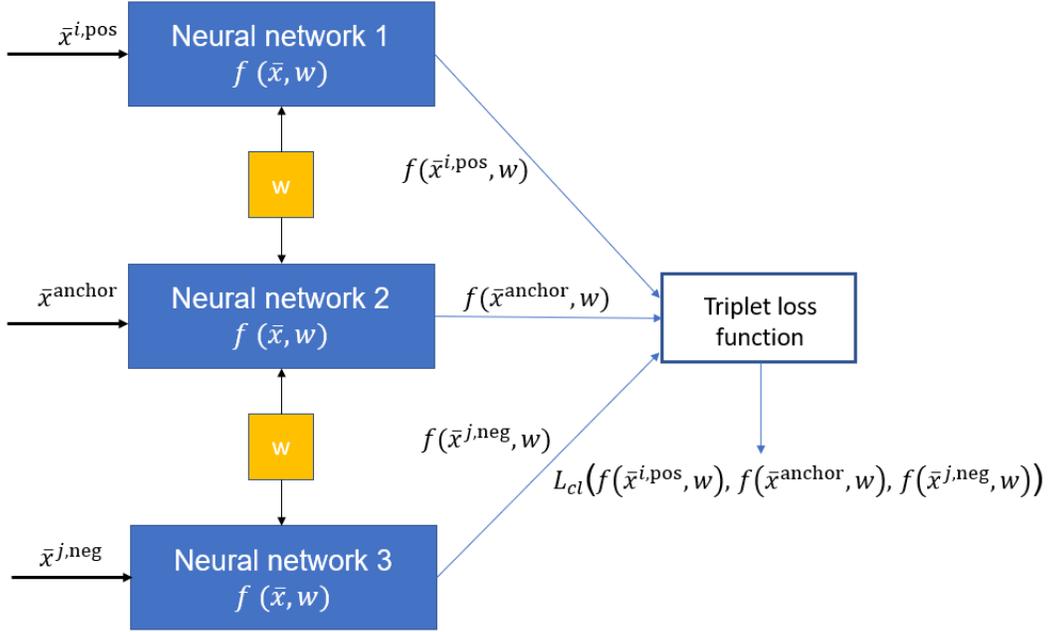}
\caption{Siamese network with triplet loss function.}
\label{fig:Unconstrained Siamese network with triplet los function}
\end{figure*}

The Siamese network with triplet loss function is
given in Figure~\ref{fig:Unconstrained Siamese network with triplet los function}
where $\bar{x}^{\text{anchor}}$ is the  input feature for a reference object
called ``anchor'', $\bar{x}^{i,\text{pos}}$ is the input feature vector for object $i$ similar to the anchor (called ``positive'' object), $\bar{x}^{j,\text{neg}}$ is the input feature vector for object $j$
dissimilar to the anchor (called ``negative'' object).
The triplet loss function is based on the following well-designed goal: 
we need to find a loss function which, in a single framework, simultaneously  minimizes the distance between similar objects (the anchor and positive objects),
and  maximizes the distance between dissimilar
objects (the anchor and negative objects). The function to achieve this goal, the triplet loss function, is defined as \cite{Schroff_et_al_1015} 
\begin{align} \label{s_triplet loss function}
& L_{tl}\big(f(\bar{x}^{i,\text{pos}},w),f(\bar{x}^{\text{anchor}},w),f(\bar{x}^{j,\text{neg}},w)\big)  \triangleq \nonumber\\   
&  \max\big(\|f(\bar{x}^{\text{anchor}},w)-f(\bar{x}^{\text{i,pos}},w)\|- \nonumber \\
& \|f(\bar{x}^{\text{anchor}},w)-f(\bar{x}^{j,\text{neg}},w)\|+m, 0\big)^2
\end{align}
where $m$ is a ``margin'' (user-defined parameter) used to help us in  pushing positive and negative objects apart.
The triplet loss function was introduced by a group of Google researchers in 2015
for the face recognition problem \cite{Schroff_et_al_1015}, and since then
it has become a popular choice for other Siamese network-based similarity learning
applications.

In the context of academic quality quantification the feature vector   $\bar{x}^{\text{anchor}}$ for the ``anchor academic'' will consist of
ideal, but realistic values; ``positive academics'' will
be those selected from the top 20 world universities based on Times Higher Education World University Rankings and QS World University Rankings, 2023; and ``negative academics''
will be those from other randomly chosen average-ranked universities, 
whose feature values $\bar{x}^{j,\text{neg}}$ are relatively weak compared to $\bar{x}^{i,\text{pos}}$ of positive academics.

\subsection{Constraints on Siamese network and architecture selection}

Again, one  structural property to be
satisfied by $f(\bar{x},w)$ in the subnetworks is that it must be a monotonically increasing function
when a feature value is replaced with a better one.
In literature there have appeared some
studies to make sure that this property holds, for example \cite{Lang_2005, Runje_Sharath_2023}.
Here, we will integrate the approaches developed in the literature for monotonic ANNs
to make sure that Siamese networks are mononotic. 
As to the architecture of the Siamese networks, they  will be  fully-connected feed-forward neural networks
since  there is no temporal or spatial relationship among the features. To enforce $f(\bar{x},w) \in [0,1]$,
normalized sigmoid function will be used in the output layer.

\section{A guideline based on the developed frameworks for academic recruitment}
\label{sec:A guideline based on the developed frameworks for academic recruitment}
\begin{figure*}[ht!]
\centering
\includegraphics[scale=0.5]{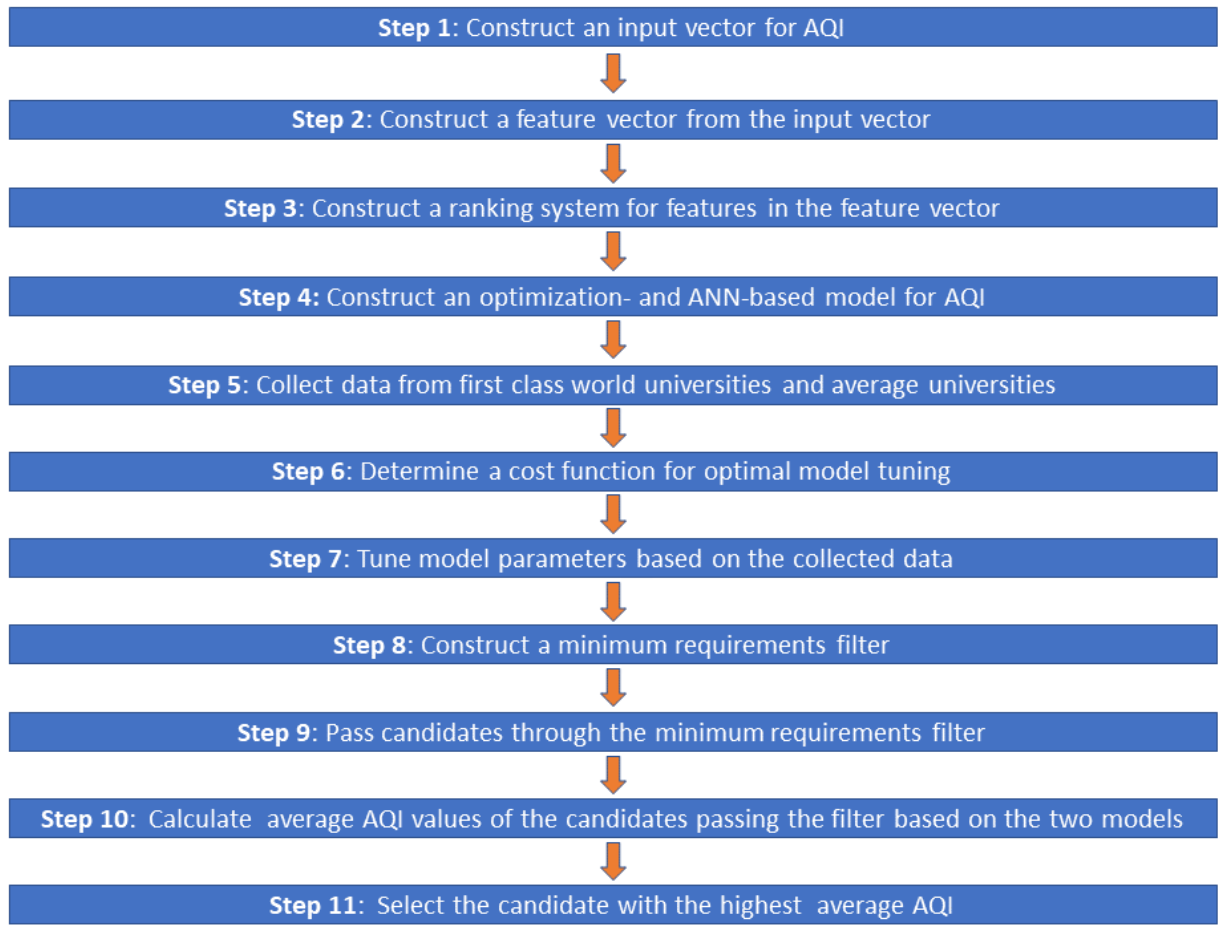}
\caption{A flowchart for model development for academic quality indexing and the use
of the resulting models.}
\label{fig:guideline_algor}
\end{figure*}

A guideline flowchart for model development for academic quality quantification and the use
of the resulting models is given in Figure~\ref{fig:guideline_algor}.
Note that a way in Step 3 for constructing a ranking system can be as follows: 
all faculty members in the related field (for example, control engineering faculty members)
can specify their feature ranking through a poll and then the
final feature ranking 
can be based on the average results.
An example of minimum requirements filter
used at Step 8 is given in Table~\ref{table:Parameters for minimum requirements}. In Table~\ref{table:Parameters for minimum requirements},
one can notice that a significant attention is paid to the national/international ranking of the university where BSc and PhD degrees
are obtained. This is very important for countries like Turkey,
Iran, China and India where students are accepted to universities based on their ranking
in national university entrance exams. In this case, the quality
of universities is extremely non-uniform.
Similar filters can constructed for a given country and field, which can vary a lot.

\begin{table}[ht]
\centering
\caption{Parameters for minimum requirements (an example filter)}
\label{table:Parameters for minimum requirements}
 \begin{tabular}{|p{1.2cm}||p{3.25cm}||p{2.5cm}|}
   \hline
   \bf{Parameter} &  \bf{Description}  & \bf{Minimum value}   \\ \hline \hline
    $p_{q_1-\text{fa}}$ & minimum  number of  SCI-SCIE papers in the ${Q_1}$ quartile as first author  & $2 \times K$ (K=1 for Assist. Prof, K=3 for Assoc. Prof, K=5 for  Prof. )  \\  \hline
    $p_\text{papers}$ & minimum  number of  total SCI-SCIE papers in the ${Q_1-Q_4}$ quartile & $2 \times L$ (L=1 for Assist. Prof, L=5 for Assoc. Prof, L=8 for  Prof. )  \\  \hline
    $p_{\text{rank-nat-BSc}}$ & minimum national ranking of university where BSc degree was obtained, if applies  & 10  \\  \hline
    $p_{\text{rank-nat-PhD}}$ &  minimum national ranking of university where PhD degree was obtained, if applies & 10   \\  \hline
    $p_{\text{rank-inat-BSc}}$ & minimum  international ranking of abroad  university where BSc degree was obtained, if applies & 100  \\  \hline
    $p_{\text{rank-inat-PhD}}$ & minimum international ranking of  abroad university where PhD degree was obtained, if applies & 100  \\  \hline
    $p_{\text{GPA}_\text{g}}$ & minimum graduate GPA  & 3.5/4.00  \\ \hline 
\end{tabular}
\end{table}

\subsection{Some caveats and remarks}
Although the presented modeling frameworks take into account many academic quality features
in calculation of AQI, there will always be  exceptional cases (outliers) for which the determined AQIs may not reflect reality. For example,
\begin{itemize}
\item There may be academics with an ``exceptional quality work'' which
can deserve prestigious awards (such as Nobel Prize), but with other academic 
quality features poor. In such cases, academic personnel selection based on the
developed approaches will miss these academics. 

\item  The research interests of some academics may involve hard theoretical
research subjects, and if additionally these subjects are not ``hot topics'' for the time being,
many of their academic quality features (such as the number of publications
and the citations they received) will be low,  which in turn will give
poor AQIs. As a result, in this scenario the quality assessment based on the presented approaches will not be realistic.

\item In such extreme cases listed above
and in other hard scenarios (including academics
with balanced theoretical and applied research interests), the AQI should be based on some other criteria, for example, totally
based on the faculty hiring committee's opinion.

\item The feature construction step can be
modified and adapted easily if it is not suitable
in its current form for some departments such as Literature
and Fine Arts departments.

\end{itemize}

Also, we want to emphasise that the objective of the
presented work is not to construct a tool using
the given approaches so that \textit{automatic} decisions
will be taken based on the resulting AQI values. Rather, 
the tool can be used as a decision-support system to help faculty hiring committee
as in other applications such as clinical
decision-support systems \cite{Musen_et_al_2014},
decision-support systems for agriculture \cite{Zhai_et_al_2020} and route planning
decision-support systems \cite{Wu_et_al_2001}.

\section{Conclusions}
\label{sec:Conclusions}

In this paper, we presented two modeling
frameworks, one based on optimization and
the other based on AI, which can be used
to develop a tool that can help as a decision-support system during evaluation of academics for faculty positions, and can help in making the evaluation process faster, transparent and fair.
The use of the developed approaches not only makes the selection of proper academics easy
given the brutal degree of competition among scholars 
in the tough academic job market, but also its use prevents
subjective assessments and stops pulling the strings.

Many academics may believe that  metrics are not the best way to assess quality in academia,
but the author of this paper has the opinion that a well-designed metric
based on an optimally balanced combination of features as done in this study is
potentially better than a subjective assessment by jury members during academic
recruitment.

In the second part, entitled ``Optimization- and AI-based approaches to academic quality quantification
for transparent academic recruitment: part 2-case studies",
a set of case studies will be presented to demonstrate 
the application of models developed in this paper.

As future research directions on this subject, academic quality quantification based on random forests or
ensemble learning approaches can be interesting directions to follow. 

\bibliographystyle{IEEEtran}

\begin{thebibliography}{50}

\bibitem{Bromley_et_al_1998}  J. Bromley, J. Bentz, L. Bottou, I. Guyon, Y. L. Cun, C. Moore, E.
Sackinger and R. Shah. ``Signature verification using a ``Siamese''  time
delay neural network'', \textit{International Journal of Pattern Recognition
and Artificial Intelligence}, vol. 7, no. 4, pp. 669-688, 1998. 

\bibitem{Annaland_et_al_2020} Y. v. S. Annaland, L. Szymanski and Steven Mills
``Predicting Cherry Quality Using Siamese Networks'',
\textit{35th International Conference on Image and Vision Computing}, 
pp.1-6, 2020.

\bibitem{Jain_et_al_2020} H. Jain, G. Harit and A. Sharma, 
``Action Quality Assessment using Siamese
Network-Based Deep Metric Learning'',  \textit{IEEE Transactions on Circuits and Systems
for Video Technology},  vol. 31, no. 6, pp. 2260-2273, 2021

\bibitem{Liu_et_al_2017} X. Liu, J. Van De Weijer and A. D. Bagdanov, ``RankIQA: Learning from Rankings for No-Reference Image Quality Assessment'',
\textit{IEEE International Conference on Computer Vision (ICCV)}, 
pp. 1040-1049, 2017.

\bibitem{Chang_et_al_2020} G. H. Chang, D. T. Felson, S. Qiu, A. Guermazi, 
T. D. Capellini and  V. B. Kolachalama, ``Assessment of knee pain from MR imaging using a convolutional Siamese network'', \textit{European Radiology}, vol. 30, pp. 3538-3548, 2020.


\bibitem{Neculoiu_et_al_2016}
P. Neculoiu, M. Versteegh and M. Rotaru, ``Learning text similarity with
siamese recurrent networks'', \textit{Proc. Represent. Workshop ACL},
pp. 148-157, 2016.

\bibitem{Wang_et_al_2018} Q. Wang, J. Gao and Y.Yuan, 
``Embedding structured contour and location
prior in siamesed fully convolutional networks for road detection''
\textit{IEEE Trans. Intell. Transp. Syst.}, vol. 19, no. 1, pp. 230-241, 2018.

\bibitem{Zhao_et_al_2019} L. Zhao, Z. Shang,  L. Zhao,  A.  Qin, and 
Y. Y. Tang, ``Siamese Dense Neural Network for Software
Defect Prediction With Small Data'', \textit{ IEEE Access}, vol. 7, pp. 7663-7677, 2019.

\bibitem{Ramachandra_et_al_2020} B. Ramachandra, M. J. Jones and R. Raju Vatsavai,
``Learning a distance function with a Siamese network to localize anomalies in videos'', 
\textit{IEEE Winter Conference on Applications of Computer Vision (WACV)},
pp. 2587-2596, 2020.

\bibitem{Koch_et_al_2015}
G. Koch, R. Zemel and R. Salakhutdinov, ``Siamese neural networks for
one-shot image recognition'', \textit{ICML Deep Learning workshop}, 2015.

\bibitem{Schroff_et_al_1015} F. Schroff, D. Kalenichenko and J. Philbin, 
``FaceNet: A unified embedding for face recognition and clustering'', 
\textit{Proc. IEEE Comput. Soc.
Conf. Comput. Vis. Pattern Recognit.}, Jun. 2015, pp. 815-823.


\bibitem{Lang_2005} B. Lang, ``Monotonic multi-layer perceptron networks as universal approximators'', \textit{Proc. Int. Conf. Artif. Neural Netw.}, pp. 31-37, 2005.

\bibitem{Runje_Sharath_2023} D. Runje and M. S. Sharath, ``Constrained Monotonic Neural Networks'', \textit{arXiv 2023}, arXiv:2205.11775, 2023.

\bibitem{Musen_et_al_2014} M. A. Musen, B. Middleton, R. A. Greenes, ``Clinical decision-support systems'' in Biomedical Informatics, New York, NY, USA:Springer, pp. 643-674, 2014.


\bibitem{Zhai_et_al_2020} Z. Zhai, J. F. Martinez, V. Beltran and N. L. Martinez, ``Decision support systems for agriculture 4.0: Survey and challenges'',
\textit{Computers and Electronics in Agriculture}, vol. 170, 105256, 2020.

\bibitem{Wu_et_al_2001} Y. H. Wu, H. J. Miller and M. C. Hung, ``A GIS based decision support system for analysis of route choice in congested urban road networks'', 
\textit{Journal of Geographical Systems}, vol. 3, no. 1, pp. 3-24, 2001.

\end{thebibliography}

\vspace{-15cm}

\begin{IEEEbiography}[{\includegraphics[width=0.9in,height=1.2in,clip]{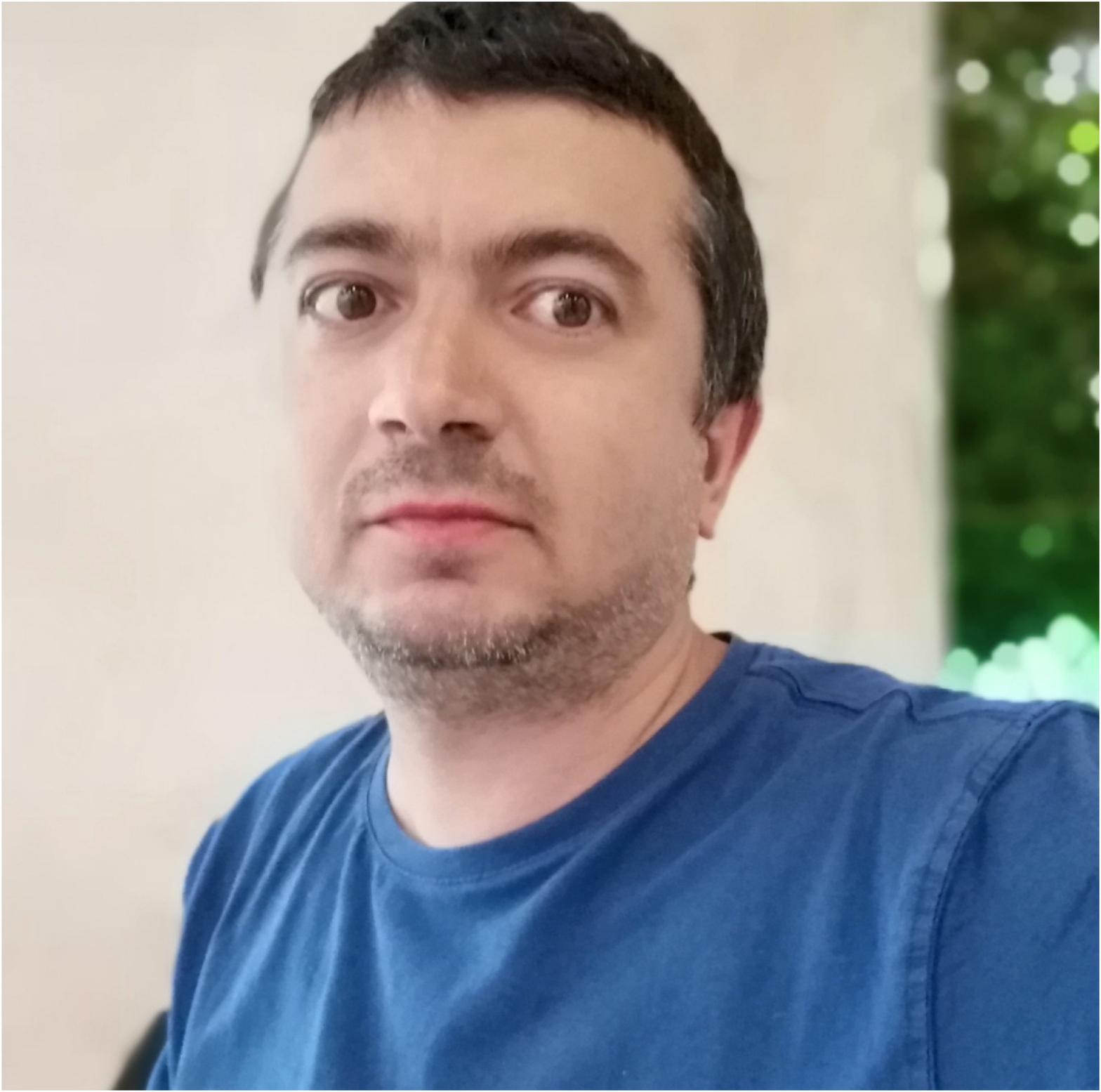}}]{\textbf{Ercan Atam}}
received his PhD  from Bo\u{g}azi\c{c}i University,
Istanbul, Turkey, in 2010. 
He was a postdoctoral
researcher with LIMSI-CNRS, France,
from 2010 to 2012,
a postdoctoral
researcher with KU Leuven, Belgium,
from 2012 to 2015 and a research associate with
Imperial College London, UK, from 2019 to 2022.
Currently, he is an associate professor at
Institute for Data Science \& Artificial Intelligence
of Bo\u{g}azi\c{c}i University, Turkey.
His interdisciplinary research interests
include control, optimization and artificial intelligence.
\end{IEEEbiography}

\end{document}